# Identifying and Decomposing Compound Ingredients in Meal Plans Using Large Language Models

**Leon Kopitar**[1,2], **Leon Bedrac**[3], **Larissa J Strath**[4,5], **Jiang Bian**[5], **Gregor Stiglic**[1,2,6]

[1]Faculty of Health Sciences, University of Maribor, Slovenia
[2]Faculty of Electrical Engineering and Computer Science, University of Maribor, Slovenia
[3]The Nu B.V., Leiden, Netherlands
[4]Pain Research & Intervention Center of Excellence (PRICE), University of Florida, Florida, USA
[5]Department of Health Outcomes and Biomedical Informatics, College of Medicine, University of Florida, Florida, USA
[6]Usher Institute, University of Edinburgh, Edinburgh, UK
{leon.kopitar1, gregor.stiglic}@um.si, leon.bedrac@gmail.com, {larissastrath, bianjiang}@ufl.edu

## Abstract

This study explores the effectiveness of Large Language Models in meal planning, focusing on their ability to identify and decompose compound ingredients. We evaluated three models—GPT-4o, Llama-3 (70b), and Mixtral (8x7b)—to assess their proficiency in recognizing and breaking down complex ingredient combinations. Preliminary results indicate that while Llama-3 (70b) and GPT-4o excels in accurate decomposition, all models encounter difficulties with identifying essential elements like seasonings and oils. Despite strong overall performance, variations in accuracy and completeness were observed across models. These findings underscore LLMs' potential to enhance personalized nutrition but highlight the need for further refinement in ingredient decomposition. Future research should address these limitations to improve nutritional recommendations and health outcomes.

## 1 Introduction

Artificial intelligence (AI) has made significant impact on various industries, including nutrition, by enhancing dietary planning and health management. With global food trends and busy lifestyles complicating meal preparation, maintaining a nutritious diet can be challenging. Nutrient profiling—a method for classifying foods based on their nutritional values—is critical for promoting healthy eating patterns and preventing illness (Alrige et al. 2017).

Effectively implementing nutrient profiling requires a thorough analysis of meal plans and individual food items, a task that can be labor-intensive and prone to errors for both healthcare providers and patients. AI improves this process by breaking down meals into their fundamental components with high accuracy. AI-driven tools offer precise quantitative and qualitative evaluations, ensuring that dietary plans are tailored to meet individual health conditions and dietary goals. This advanced approach not only allows for more personalized meal adjustments but also highlights the growing need for sophisticated, AI-powered solutions to analyze and optimize the nutritional profiles of diverse food combinations, supporting personalized nutrition and overall health.

Large Language Models (LLMs) excel in interpreting and producing text that closely resembles human communication. With training on extensive datasets, these models can understand and analyze intricate concepts, making them well-suited for a range of text processing tasks and advancing the pursuit of artificial general intelligence (AGI) in multiple fields (Wei et al. 2022; Naveed et al. 2023). The semantic and contextual abilities of LLMs make it possible to create systems that analyze meal plans and offer customized nutritional recommendations (Ma et al. 2024; Lu et al. 2020).

Meal plans often feature foods in their completed form, such as a lasagna, without breaking them down into their component ingredients. Ingredients fall into two categories: basic, which are singular and cannot be further decomposed, and compound, which consist of multiple basic ingredients combined together.

Therefore, decomposing compound ingredients within meal plans is crucial for several reasons. First, allergy and intolerance management is enhanced by identifying the individual components of compound ingredients. This allows for the substitution of problematic elements, ensuring meals are safe for those with specific dietary restrictions. Second, nutritional accuracy is improved by analyzing each ingredient separately, which allows for a precise assessment of the meal's overall nutritional profile and ensures it meets dietary guidelines and health standards. Third, cost estimation becomes more effective by breaking down ingredients, facilitating budget planning and cost management, especially when preparing meals in large quantities or for institutional settings. Next, personalization of meal plans is better supported when ingredients are decomposed, enabling customization to meet individual dietary preferences, health goals, and nutritional needs. And final, nutritional compliance is achieved by aligning meal plans with food group guidelines and using structured frameworks to monitor and adjust nutrient levels. This approach helps prevent both excesses of harmful nutrients and deficiencies of essential nutrients, contributing to healthier dietary patterns and supporting public health initiatives.

In a study, a smartphone-based application addresses the challenge of meal composition by estimating the energy and macronutrient content (carbohydrates, proteins, and fats) from user-captured food images. The system utilizes deep neural networks, specifically employing a segmentation method within the Mask R-CNN framework (Abdulla 2017), to identify, segment, and analyze the foods depicted in the images (Lu et al. 2020).

This paper explores the effectiveness of LLMs in identifying and breaking down compound ingredients into their basic components and aligning these components with entries in the USDA FoodData Central repository. Our approach relies solely on the meal's name and description to assess how LLMs can be used effectively even in the absence of detailed information. Although this study establishes a preliminary understanding of LLMs' capabilities in ingredient decomposition and knowledge representation (KR), it represents just the beginning of an ongoing research effort. Future work will build on these findings by incorporating detailed nutritional estimations and exploring additional applications of KR, with the aim of enhancing the precision and personalization of dietary recommendations.

## 2 Materials and Methods

### 2.1 Study Design and Research Approach

We assessed the performance of three LLMs: GPT-4o (OpenAI 2024a), Llama-3 (70b) (Meta 2024), and Mixtral (8x7b) (Jiang et al. 2024). The evaluation focused on each model's ability to identify and decompose compound ingredients into basic components. To verify the accuracy of the results, the decomposed ingredients were cross-referenced with entries in the USDA FoodData Central repository.

The research employed programming language Python and key libraries such as OpenAI (OpenAI 2024b) and Perplexity (Perplexity AI 2024). GPT-4o was utilized to generate meal plans and map the identified basic ingredients to the USDA FoodData through its API (USDA 2019). These libraries were instrumental in both recognizing and deconstructing compound ingredients.

Figure 1 displays a flowchart of the entire research process, which includes the generation of meal plans, the identification and decomposition of compound ingredients, and evaluation. Comprehensive quality control will be conducted in the later stages of the research and is not included in the current analysis.

### 2.2 Generation of Meal Plans

With GPT-4o's default settings, 15 meal plans were generated in five iterations, including five each for breakfast, lunch, and dinner. Each plan specified exact portion sizes and ingredient quantities, with the plans created independently.

### 2.3 Ingredient Identification and Decomposition

For each model (GPT-4o, Llama-3 (70b), Mixtral (8x7b)), we analyzed 15 meal plans and tasked the models with identifying and deconstructing compound ingredients into their simpler components, basic ingredients. For instance, 'Beef Stroganoff' is a compound dish that can be deconstructed into basic ingredients like beef strips, sour cream, onions, mushrooms, beef broth, flour, and butter. These fundamental components were matched against the USDA FoodData entries, and their nutritional content was calculated and aggregated for the whole dish.

### 2.4 Model Evaluation and Performance Metrics

The evaluation of LLMs centers on their ability to accurately identify and decompose compound ingredients into their fundamental components. To establish a benchmark, nutritionists review 15 meal plans to identify the compound ingredients, which are then considered the ground truth for the assessment. These expert-identified ingredients are compared with those predicted by the models to assess their performance. The comparison is used to calculate key metrics such as accuracy and the F1-score. Accuracy measures the proportion of correctly identified compound ingredients relative to the total number assessed, while the F1-score provides a balanced evaluation of precision and recall, which is especially important in cases of class imbalance. Additionally, nutritionists review the decomposed basic ingredients and their quantities to ensure that they are realistic and consistent with the original compound ingredients.

The statistical significance of performance differences between models is assessed using either a paired t-test or a Wilcoxon signed-rank test. The choice of test depends on the data's distribution, as determined by the Shapiro-Wilk test.

## 3 Results

In this section, we present the outcomes of our evaluation across three models—Llama-3 (70b), GPT-4o, and Mixtral (8x7b)—focusing on their ability to identify and decompose compound ingredients within meal plans. The performance metrics and comparative analysis reveal each model's strengths and weaknesses in accurately handling ingredient data. We first address the identification of compound ingredients, followed by an examination of how effectively each model decomposes these ingredients into their basic components. The findings provide a comprehensive overview of the models' capabilities and highlight key areas for improvement.

### 3.1 Identification of Compound Ingredients

In the analysis of 15 meal plans, 101 ingredients were identified, reduced to 99 unique ingredients after accounting for repetitions. Figure 2 illustrates the accuracy and F1-score for each model, including their mean values and 95% confidence intervals. The analysis shows that Llama-3 (70b) and GPT-4o exhibit no statistically significant difference in performance for identifying compound ingredients ($p > .1$). Llama-3 (70b) had an F1-score of 0.894 (95% CI: 0.84-0.95) and an accuracy of 0.893 (95% CI: 0.85-0.94), while GPT-4o had an F1-score of 0.842 (95% CI: 0.79-0.89) and an accuracy of 0.835 (95% CI: 0.78-0.89), while Mixtral (8x7b) performance is significantly worse than both models in both metrics ($p < .01$).

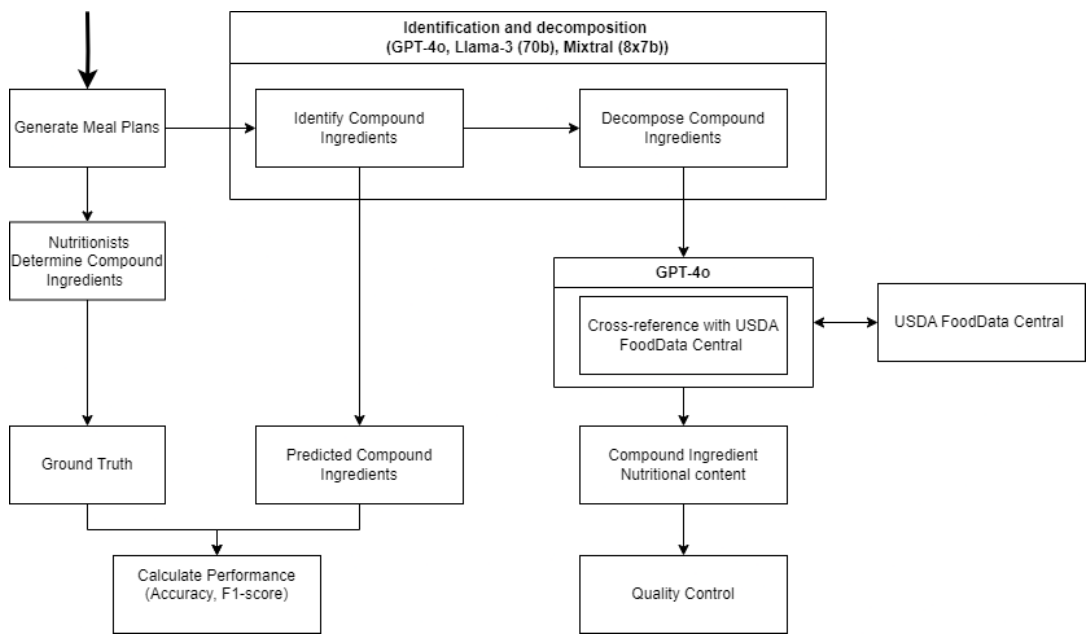

Figure 1: Flowchart illustrating the research process.

## 3.2 Decomposition of Compound Ingredients

The models exhibited different frequencies and accuracies in decomposing compound ingredients. GPT-4o and Llama-3 (70b) included basic ingredients such as salt, sugar, or pepper less frequently (3-times and 2-times) compared to Mixtral (8x7b), which included these ingredients more often (8-times).

Evaluation of the GPT-4o, Llama-3 (70b), and Mixtral (8x7b) models in decomposing compound ingredients showed varying accuracy across different meals. Common problems included missing oils, seasonings, and sweeteners in dishes like scrambled eggs and mixed salads. GPT-4o often failed to decompose items like whole wheat toast, while Llama-3 (70b) showed slightly better performance, such as including olive oil in ratatouille. Mixtral (8x7b) had difficulties with more complex items like hummus but managed simpler ones accurately. All models struggled with integrating seasonings and cooking methods, leading to incomplete nutritional profiles.

## 4 Discussion

In this study, we evaluated three models—Llama-3 (70b), GPT-4o, and Mixtral (8x7b)—on their ability to identify and decompose compound ingredients within meal plans. The results reveal clear distinctions among the models in their performance, with significant implications for future improvements in ingredient decomposition and knowledge representation.

Overall, Mixtral (8x7b) consistently fell short of GPT-4o and Llama-3 (70b) in accuracy and F1-score, revealing significant gaps in several critical aspects. In contrast, GPT-4o and Llama-3 (70b) demonstrated comparable performance levels, with no statistically significant differences found between the two.

Although all models demonstrated some ability to identify basic ingredients, none consistently included crucial elements such as salt, pepper, and sugar with sufficient frequency. However, Mixtral (8x7b) incorporated these basic ingredients more often than GPT-4o and Llama-3 (70b), with a rate of 12.9% compared to 9.7% for GPT-4o and 6.5% for Llama-3 (70b). This indicates that Mixtral (8x7b) tends to prioritize these essential ingredients more reliably when generating recipes or related content.

The analysis reveals both strengths and weaknesses in the GPT-4o, Llama-3 (70b), and Mixtral (8x7b) models regarding meal decomposition. Although all models demonstrate some capability in identifying basic ingredients, they struggle with consistently handling oils, seasonings, and compound ingredients, highlighting areas for improvement. Llama-3 (70b) excels in certain culinary details, suggesting it may be better suited for applications that require detailed ingredient decomposition. Nevertheless, the frequent omission of essential ingredients by all models underscores the need for improved algorithms for more comprehensive ingredient recognition. Future enhancements should aim to address these issues, providing more accurate and reliable nutritional analysis to help users better understand the nutritional content of their meals.

The findings from this study have substantial real-world implications. For example, incorporating ingredient decomposition into dietary planning tools could improve the accuracy and personalization of nutrition recommendations, making them better suited to individual dietary requirements. Such tools would benefit from the model's capability to provide precise ingredient quantities and compositions, which are essential for creating detailed nutritional profiles. Furthermore, this research could enhance health apps that offer dietary advice, improving their effectiveness in managing specific health conditions or reaching dietary objectives. Future exploration of these applications could bridge the gap between theoretical insights and practical implementation, delivering significant benefits to users.

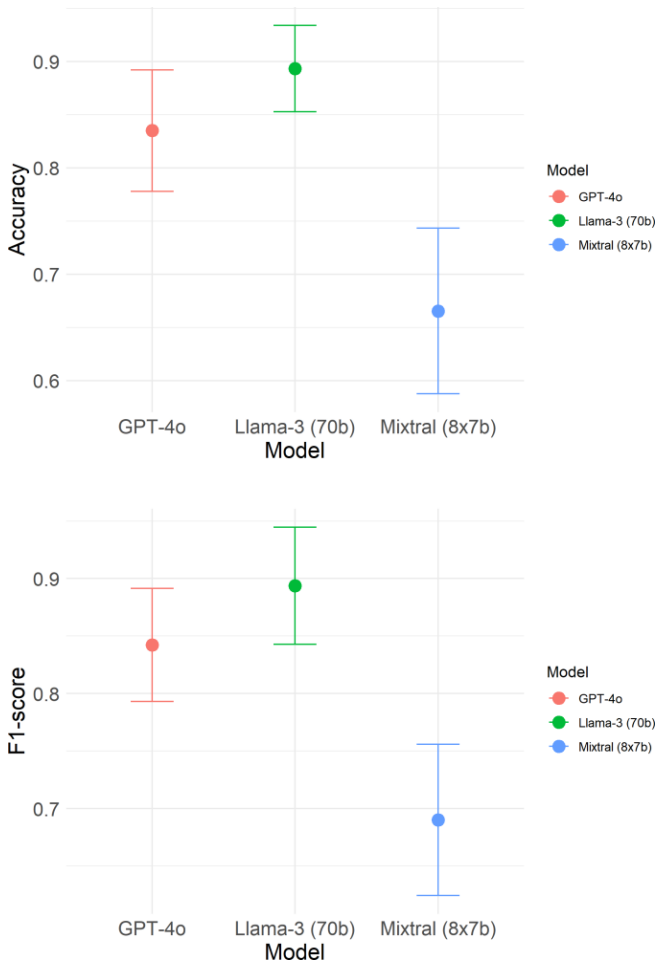

Figure 2: Prediction performance (Accuracy and F1-score) with 95% confidence intervals.

Effective knowledge representation (KR) is crucial for understanding and manipulating complex data, such as ingredient lists in meal plans. It involves structuring and formalizing information in ways that enable models to process and interpret data effectively. By accurately identifying and classifying ingredients, KR enhances various applications, including personalized nutrition, diet planning, and food science research (Theodore Armand et al. 2024). For instance, representing a 'Grilled Vegetable Frittata' not just as a compound ingredient but as a combination of basic ingredients (e.g., eggs, bell peppers, onions) allows for more precise nutritional analysis, dietary replacements, and recommendations.

Furthermore, KR aids in creating personalized meal plans that adhere to dietary guidelines through rules and ontology implementation. For instance, it helps to ensure macronutrients are balanced, with proteins at 10-20%, carbohydrates at 45-60%, and fats at 25-40% of total energy intake. It also helps limit saturated fats to under 10% and sugar to under 10% of total calories (Vepsäläinen and Lindström 2024; Blomhoff et al. 2023), or checking whether we are reaching the 14 grams of fiber per 1000 calories as suggested in the guidelines (U.S. Department of Agriculture 2020).

KR ensures that meal plans can be adjusted based on different goals. For example, three meals a day with an optional fourth for muscle gain or two meals for time-restricted feeding, where sufficient protein intake is the key, with targets of $\approx 1.6$ grams per kilogram of body weight per day, with potentially higher requirements for those over 40 to aid muscle synthesis (Schoenfeld and Aragon 2018; Morton et al. 2018; Moore 2021; Trommelen, Betz, and van Loon 2019). KR's techniques like ontologies or semantic networks organize this data to ensure accuracy and relevance in dietary recommendations.

## 5 Conclusion

Our study highlights the capabilities of LLMs in significantly enhancing meal planning by accurately identifying compound ingredients. These models demonstrate a strong proficiency in recognizing complex ingredient combinations, contributing to more efficient and informed meal preparation. However, despite their strengths, LLMs fall short in the precise decomposition of compound ingredients into their individual components. This limitation suggests that while LLMs can greatly enhance meal planning, there remains room for improvement in their ability to break down and analyze compound ingredient content.

Preliminary results highlight the promise of LLMs in improving meal planning through effective identification and interpretation of complex ingredient combinations. Models such as GPT-4o, Llama-3 (70b), and Mixtral (8x7b) demonstrate notable skills in recognizing compound ingredients, though performance varies significantly among them.

## Acknowledgements

This project is funded by the European Union's Horizon Europe research and innovation programme under the Grant Agreement No 101159018. Views and opinions expressed are however those of the author(s) only and do not necessarily reflect those of the European Union or the European Research Executive Agency (REA). Neither the European Union nor the granting authority can be held responsible for them.

This work was supported in part by the Slovenian Research Agency under Grant ARRS P2-0057.

## References

Abdulla, W. 2017. Mask r-cnn for object detection and instance segmentation on keras and tensorflow. https://github.com/matterport/Mask_RCNN.

Alrige, M. A.; Chatterjee, S.; Medina, E.; and Nuval, J. 2017. Applying the concept of nutrient-profiling to promote healthy eating and raise individuals' awareness of the nutritional quality of their food. In *AMIA Annual Symposium Proceedings*, volume 2017, 393. American Medical Informatics Association.

Blomhoff, R.; Andersen, R.; Arnesen, E. K.; Christensen, J. J.; Eneroth, H.; Erkkola, M.; Gudanaviciene, I.; Halldórsson, . I.; Ho¨yer-Lund, A.; Lemming, E. W.; et al. 2023. *Nordic Nutrition Recommendations 2023: integrating environmental aspects*. Nordic Council of Ministers.

Jiang, A. Q.; Sablayrolles, A.; Roux, A.; Mensch, A.; Savary, B.; Bamford, C.; Chaplot, D. S.; de las Casas, D.; Hanna, E. B.; Bressand, F.; Lengyel, G.; Bour, G.; Lample, G.; Lavaud, L. R.; Saulnier, L.; Lachaux, M.-A.; Stock, P.; Subramanian, S.; Yang, S.; Antoniak, S.; Scao, T. L.; Gervet, T.; Lavril, T.; Wang, T.; Lacroix, T.; and Sayed, W. E. 2024. Mixtral of Experts.

Lu, Y.; Stathopoulou, T.; Vasiloglou, M. F.; Pinault, L. F.; Kiley, C.; Spanakis, E. K.; and Mougiakakou, S. 2020. gofoodtm: an artificial intelligence system for dietary assessment. *Sensors* 20(15):4283.

Ma, P.; Tsai, S.; He, Y.; Jia, X.; Zhen, D.; Yu, N.; Wang, Q.; Ahuja, J. K.; and Wei, C.-I. 2024. Large language models in food science: Innovations, applications, and future. *Trends in Food Science & Technology* 104488.

Meta. 2024. Introducing Meta Llama 3: The Most Capable Openly Available LLM to Date. Accessed: 2024-06-29.

Moore, D. R. 2021. Protein requirements for master athletes: just older versions of their younger selves. *Sports Medicine* 51(Suppl 1):13–30.

Morton, R. W.; Murphy, K. T.; McKellar, S. R.; Schoenfeld, B. J.; Henselmans, M.; Helms, E.; Aragon, A. A.; Devries, M. C.; Banfield, L.; Krieger, J. W.; et al. 2018. A systematic review, meta-analysis and meta-regression of the effect of protein supplementation on resistance training-induced gains in muscle mass and strength in healthy adults. *British journal of sports medicine* 52(6):376–384.

Naveed, H.; Khan, A. U.; Qiu, S.; Saqib, M.; Anwar, S.; Usman, M.; Barnes, N.; and Mian, A. 2023. A comprehensive overview of large language models. *arXiv preprint arXiv:2307.06435*.

OpenAI. 2024a. Hello GPT-4o — OpenAI. Accessed: 2024-06-29.

OpenAI. 2024b. Openai API. Accessed: 2024-07-03.

Perplexity AI. 2024. Perplexity API. Accessed: 2024-07-03.

Schoenfeld, B. J., and Aragon, A. A. 2018. How much protein can the body use in a single meal for muscle-building? implications for daily protein distribution. *Journal of the International Society of Sports Nutrition* 15:1–6.

Theodore Armand, T. P.; Nfor, K. A.; Kim, J.-I.; and Kim, H.-C. 2024. Applications of artificial intelligence, machine learning, and deep learning in nutrition: A systematic review. *Nutrients* 16(7):1073.

Trommelen, J.; Betz, M. W.; and van Loon, L. J. 2019. The muscle protein synthetic response to meal ingestion following resistance-type exercise. *Sports Medicine* 49(2):185–197.

U.S. Department of Agriculture. 2020. Dietary Guidelines for Americans 2020 - 2025 : Make Every Bite Count With the Dietary Guidelines. *The American journal of clinical nutrition* 34(1):121–123.

USDA. 2019. U.S. Department of Agriculture, Agricultural Research Service FoodData Central. Accessed: 2024-06-29.

Vepsa¨la¨inen, H., and Lindstro¨m, J. 2024. Dietary patterns – a scoping review for Nordic Nutrition Recommendations 2023. *Food and Nutrition Research* 68.

Wei, J.; Wang, X.; Schuurmans, D.; Bosma, M.; Xia, F.; Chi, E.; Le, Q. V.; Zhou, D.; et al. 2022. Chain-of-thought prompting elicits reasoning in large language models. *Advances in neural information processing systems* 35:24824–24837.